\documentclass[sigconf]{acmart}
\renewcommand\footnotetextcopyrightpermission[1]{}
\AtBeginDocument{%
  }
\usepackage{multirow}

\setcopyright{acmlicensed}
\copyrightyear{2026}
\acmYear{2026}
\acmDOI{XXXXXXX.XXXXXXX}

\newcommand{\projecttitle}{TrioMan}
\newcommand{\myparagraph}[1]{\smallskip \noindent{\bf {#1}.}}

\begin{document}

\title{Generator–Refiner–Examiner: \\ A Tri-Module Data Augmentation Framework for 3D Human Avatar Learning from Monocular Videos}

\author{Gangjian Zhang}
\affiliation{%
  \institution{HKUST(GZ)}
  \city{Guangzhou}
  \country{China}
}

\author{Jian Shu}
\affiliation{%
  \institution{HKUST(GZ)}
  \city{Guangzhou}
  \country{China}
}

\author{Sicheng Yu}
\affiliation{%
  \institution{HKUST(GZ)}
  \city{Guangzhou}
  \country{China}
}

\author{Wenhao Shen}
\affiliation{%
  \institution{Nanyang Technological University}
  \country{Singapore}
}

\author{Yu Feng}
\affiliation{%
  \institution{HKUST(GZ)}
  \city{Guangzhou}
  \country{China}
}

\author{Hao Wang}
\authornote{Corresponding author. \\
Email: \texttt{\{gzhang292, haowang\}@connect.hkust-gz.edu.cn} \\
Demo page: \url{https://gre2026.github.io/GRE2026/}
}
\affiliation{%
  \institution{HKUST(GZ)}
  \city{Guangzhou}
  \country{China}
}


\begin{abstract}
This paper addresses the challenge of reconstructing photorealistic and animatable 3D human avatars from monocular videos. While existing methods rely on combining per-subject optimization with generic human priors, they often fail to capture fine-grained details when training frames are limited. To mitigate this data scarcity, we propose TrioMan, a systematic tri-module framework for augmented 3D avatar learning. Our approach comprises three synergistic components. The Generator creates diverse unseen samples by imposing Gaussian perturbations on pose and camera. The Refiner improves the quality of generated data through one-step diffusion guided by texture and geometry cues. The Examiner selects subject-consistent samples using a dual-branch attention-based similarity evaluation. Experiments on the X-Humans and NeuMan benchmarks show that TrioMan outperforms state-of-the-art methods. 
\end{abstract}



\begin{teaserfigure}
  \includegraphics[width=\textwidth]{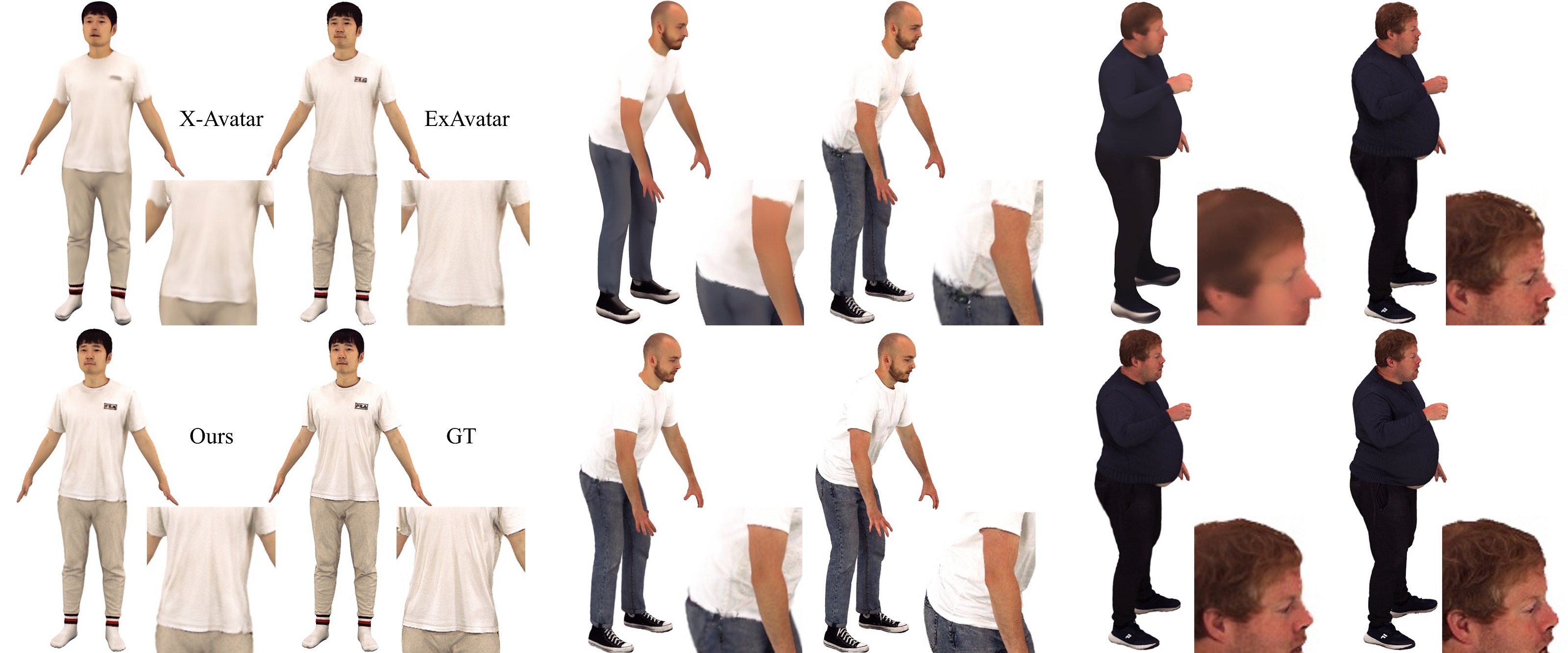}
  \caption{\textbf{Qualitative Comparison.} We use the same SMPL templates to drive the animatable 3D avatars of different SOTA methods and render 2D images. In comparison, our method achieves more accurate reconstruction of clothing wrinkles, smoother clothing connections, and fewer visual artifacts. Please refer to the supplementary material for more video demos.}
  \label{fig:tease}
\end{teaserfigure}


\maketitle

\section{Introduction}
\label{sec:intro}

Photorealistic, animatable 3D avatars are essential for immersive technologies like VR and AR. Reconstructing high-fidelity avatars from monocular videos is a key goal, as it alleviates the need for costly multi-view capture systems~\cite{multiview-sys, chen2024meshavatar} or 3D scans. Recent progress in 3D Gaussian Splatting (3DGS)~\cite{3DGaussian} has boosted avatar reconstruction from monocular input~\cite{moon2024exavatar, hu2024expressive_eva, hu2024gaussianavatar, guo2023vid2avatar}.

However, monocular video-based reconstruction~\cite{moon2024exavatar, hu2024expressive_eva, peng2025parametric, guo2025vid2avatar, 3DGS-Avatar, hu2024gaussianavatar, guo2023vid2avatar, jiang2022neuman, jiang2022instantavatar, weng_humannerf_2022_cvpr} still suffers from limited input diversity and insufficient viewpoint coverage. A typical monocular video~\cite{shen2023xavatar, jiang2022neuman} lasts less than a minute, capturing only a small set of poses and camera angles. This narrow range restricts the model’s ability to reconstruct consistent geometry and appearance under unseen motions or views.

To improve reconstruction performance under limited data, recent works such as PGHM~\cite{peng2025parametric} and Vid2Avatar-Pro~\cite{guo2025vid2avatar} introduce prior knowledge. The former uses multi-view human datasets~\cite{cheng2023dna, xiong2024mvhumannet} to train parametric priors that regularize reconstruction and reduce overfitting. The latter learns a universal prior from 1,000 in-the-wild performances and fine-tunes it on the target subject. While these priors improve pose generalization, they focus on generic human characteristics than more pertinent subject-specific details. As a result, they struggle to directly address the fundamental limitation of capturing personalized dynamics from monocular input.

Toward this end, this paper introduces \projecttitle, a tri-module augmented avatar learning framework designed to enrich training diversity, enhance synthetic-real consistency, and ensure sample reliability.
Built upon 3DGS~\cite{3DGaussian} and SMPL-X~\cite{SMPL-X:2019} modeling, \projecttitle~features three tightly coupled modules: the Generator, Refiner, and Examiner. 

Collectively, these components address the challenge of limited frame data in monocular video setups, without compromising the fine-grained, subject-specific details vital for realistic avatar reconstruction.

To be specific, 
(1) the Generator synthesizes diverse unseen training frames by sampling pose and camera perturbations from Gaussian distributions. It improves the data abundance beyond the range of the original video, addressing the limitation of training data scarcity. 
(2) The Refiner uses one-step diffusion~\cite{yin2024one, wu2025difix3d} with texture and geometry conditioning to enhance the Generator’s coarse frames into photorealistic images. It improves the consistency between synthetic and real samples.
(3) The Examiner acts as a quality gatekeeper, evaluating the similarity between refined frames and real video frames using a dual-branch attention-based network. It ensures that only high-fidelity, subject-consistent pseudo samples are used for training. 
By integrating these three modules, \projecttitle~equips the model with rich and realistic training samples, covering novel human poses and camera views.

 Experiments on benchmark datasets (X-Humans, NeuMan) validate that our proposed method achieves SOTA performance.  Our main contributions are threefold: 

\begin{itemize}

\item We propose a tri-module augmented avatar learning framework that systematically addresses monocular data scarcity by integrating data synthesis, synthetic-real alignment, and quality control in a unified pipeline.

\item We design a distribution-perturbed Generator and a one-step diffusion Refiner, which work collectively to produce diverse, photorealistic training samples while retaining subject-specific characteristics.

\item We introduce a dual-branch attention-based Examiner that mitigates the stochasticity of diffusion-based generation and provides reliable pseudo-GT for model training.

\end{itemize}

\section{Related Work}
\label{sec:rw}
\subsection{Monocular Video Human Reconstruction}

Reconstructing 3D human avatars from monocular videos is a critical task in computer vision, with applications in VR/AR and more. The rise of Neural Radiance Fields (NeRF) \cite{mildenhall2020nerf} enabled implicit articulated avatar reconstruction from monocular video \cite{nerf1, nerf2, nerf3, nerf4, nerf5, nerf6, jiang2022neuman, guo2023vid2avatar, jiang2022instantavatar, yu2023monohuman}. Recent advances have shifted to 3D Gaussian Splatting (3DGS)~\cite{3DGaussian} for faster rendering and simpler optimization \cite{opim_gau1, opim_gau2, opim_gau3, opim_gau4}. Two paradigms have emerged: direct per-subject Gaussian parameter optimization \cite{opim_gau1, opim_gau2} and neural network-based attribute prediction \cite{hu2024gaussianavatar_neunetpred, hu2024gauhuman, kocabas2024hugs_neunetpred, 3DGS-Avatar, li2023human101training100fpshuman, liu2025vga, wen2024gomavatar_neunetpred, hu2024expressive_eva}. GaussianAvatar~\cite{hu2024gaussianavatar_neunetpred} uses SMPL’s UV map for pose-dependent effects, while ExAvatar \cite{exavatar_neunetpred} integrates SMPL-X for whole-body (face/hand) control. EVA~\cite{hu2024expressive_eva} enhances avatar expressiveness in detailed areas through an SMPL-X alignment module and adaptive density control. More recently, Vid2Avatar-Pro \cite{guo2025vid2avatar} creates high-quality animatable avatars from monocular videos using a universal prior and diffusion-based texture inpainting. PGHM \cite{peng2025parametric} combines parametric human priors with 3D Gaussian Splatting for fast and high-fidelity avatar reconstruction. PriorAvatar \cite{PriorAvatar} utilizes a multi-person feature codebook as a 3D human prior to guide Gaussian-based avatar fitting. MonoCloth \cite{jin2026monocloth} employs a part-based decomposition strategy and a cloth simulation module to improve reconstruction and animation realism.



\subsection{3DGS for Human Representation}


The efficiency and high quality of 3DGS have led to its rapid adoption for realistic human avatar representation. Early works~\cite{hugs2024human, liu2024humangaussian, sim2025persona} established frameworks that map canonical Gaussians to target poses via deformation models, while hybrid representations~\cite{shao2024splattingavatar, moon2024exavatar}, where Gaussians are explicitly embedded onto a deformable mesh for enhanced structural fidelity, have been developed. Beyond subject-specific reconstruction, other research has further leveraged 3DGS for efficient text-driven 3D human generation through structure-aware sampling~\cite{GPS-Gaussian, kocabas2024hugs, liu2024humangaussian}. Recently, large Gaussian models for human avatar reconstruction have been widely explored and have demonstrated outstanding texture details~\cite{zhang2025multigo, qiu2025lhm, qiu2025pf, yao2026multigo++,zhang2025sat, shu2025fastanimatelearnabletemplateconstruction}, enabling the synthesis of Gaussians in a feed-forward approach.

\begin{figure*}
    \centering
    \includegraphics[width=1.0\linewidth]{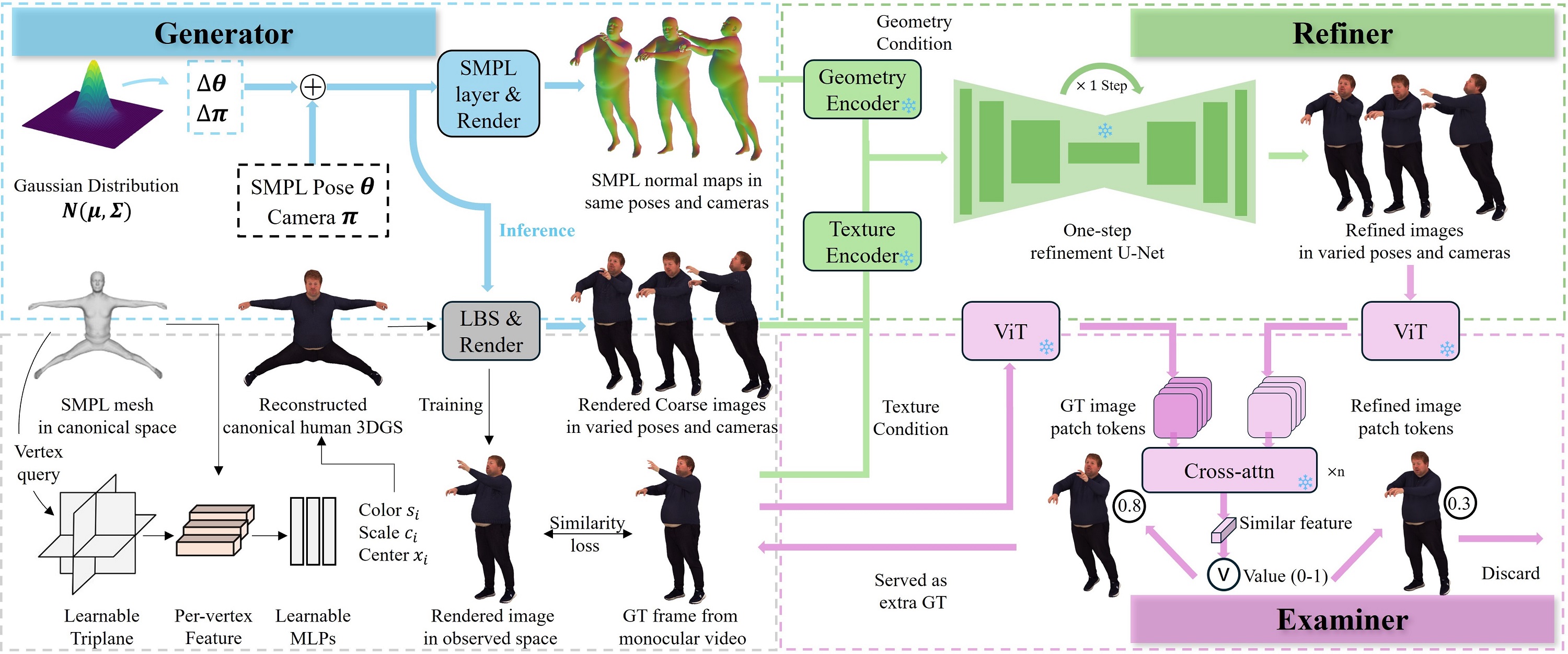}
    \caption{\textbf{Method Overview.} Our method, \projecttitle, addresses expressive 3D human avatar learning from monocular video by introducing a tri-module (Generator-Refiner-Examiner) augmented avatar learning framework. The Generator leverages Gaussian distributions to sample pose and camera perturbations, which are fused with the SMPL pose (fitted from the video frame) and camera pose to generate SMPL pose variations that can drive the Human 3D Gaussian reconstructed from the baseline model to produce unseen coarse frames. The Refiner takes geometric conditions from variation SMPLs and texture conditions from the real frame, refining them via a one-step diffusion process to generate photorealistic refined frames. Moreover, the Examiner assesses the similarity in details between the refined frame and the real video frame to determine whether the refined frame will be included as a pseudo ground truth (GT) for the avatar learning. } 
    \label{fig: pipeline}
\end{figure*}


\subsection{3DGS for Scene Representation}

Extending 3DGS to dynamic scenes has attracted significant attention \cite{yu2024cogs, huang2024sc-gs, li2024spacetime}. Mainstream approaches often involve learning complex deformation fields \cite{wu20244dgs, yang2024deformable-gs} or directly optimizing Gaussian attributes across time \cite{yang2023realtime, duan20244d-rotor, cheng2025graphguidedscenereconstructionimages}. While recent methods combining triplane and hash-coding \cite{xu2024grid4d} or spline functions \cite{park2025splinegs} show impressive results, they often lack smoothness or require massive priors, such as 2D trajectories and depth estimation. However, these methods typically assume dense multi-view supervision or known camera poses, which are unavailable in monocular videos. Furthermore, they rarely model articulated human topology, leading to distorted limbs during fast motion. While recent works \cite{wu2025difix3d, liu2024reconx, lin2024gaussianflow, zhong2025taming, yu2024viewcrafter} leverage generative priors, like diffusion models, to refine details or regularize reconstructions from sparse inputs, their application is limited to static refinement and does not address the core data scarcity of novel poses and viewpoints in monocular human video.


\section{Methodology}

\subsection{Preliminaries}
\label{prel}
\myparagraph{Gaussian Splatting} Gaussian splatting~\cite{3DGaussian}, emerging as a popular 3D representation, utilize a collection of 3D Gaussians, denoted by $\mathcal{G}$, to model 3D data. Each 3D Gaussian is characterized by a parameter set: $\{ \mathbf{x}_i, \mathbf{s}_i, \mathbf{q}_i, \alpha_i, \mathbf{c}_i \} \in \mathbb{R}^{14}$. Here, $\mathbf{x} \in \mathbb{R}^3$ represents the geometry center, $\mathbf{s} \in \mathbb{R}^3$ the scaling factor, $\mathbf{q} \in \mathbb{R}^4$ the rotation quaternion, $\alpha \in \mathbb{R}$ the opacity value, and $\mathbf{c} \in \mathbb{R}^3$ the color feature.

\myparagraph{SMPL(X) $\&$ LBS} Skinned Multi-Person Linear (SMPL) model \cite{loper2023smpl} is a parametric human body model that generates 3D meshes by linearly blending shape and pose variations. It maps low-dimensional body parameters (Shape $\beta$ and Pose $\theta$) to 3D mesh, facilitating efficient reconstruction and animation. The human body mesh $\mathcal{M}$ is defined as: $\mathcal{M}(\beta, \theta): (\beta, \theta) \Rightarrow \mathbb{R}^{3\times6890}$. Extensions like SMPL-X \cite{SMPL-X:2019, romero2022embodied_smplh} enhance control over facial expressions, body movements. 

Linear Blend Skinning (LBS) is a mesh deformation method used in models like SMPL, deforming the template mesh using parameters ($\theta$) for body articulation.


\myparagraph{Diffusion Model (DM)} DM~\cite{ho2020denoisingdiffusionprobabilisticmodels, lin2024sdxl, SDTurbo} learns to model the data distribution $p_{\text {data }}(\mathbf{x})$ through iterative denoising and are trained with denoising score matching. Specifically, to train a diffusion model, diffused versions $\mathbf{x}_\tau= \mathbf{x}+\sigma_\tau \boldsymbol{\epsilon}$ of the data $\mathbf{x} \sim p_{\text {data }}$ are generated, by progressively adding Gaussian noise $\epsilon \sim \mathcal{N}(\mathbf{0}, \boldsymbol{I})$. Learnable parameters $\boldsymbol{\theta'}$ of the denoiser model $\mathbf{F}_{\boldsymbol{\theta'}}$ are optimized via the objective:
\begin{equation}
\mathbb{E}_{\mathbf{x} \sim p_{\text {data }}, \tau \sim p_\tau, \boldsymbol{\epsilon} \sim \mathcal{N}(\mathbf{0}, \boldsymbol{I})}\left[\left\|\mathbf{y}-\mathbf{F}_{\theta'}\left(\mathbf{x}_\tau ; \mathbf{c}, \tau\right)\right\|_2^2\right],
\end{equation}
where $\mathbf{c}$ is condition. The target vector $\mathbf{y}$ is usually set as the added noise $\epsilon$. Finally, $p_\tau$ denotes a uniform distribution over the diffusion time variable $\tau$. 


\subsection{The Proposed Method}
\subsubsection{\textbf{Overview}}


Our method, \projecttitle, outlined in Figure~\ref{fig: pipeline}, enhances the learning of expressive 3D human avatars from monocular RGB videos through a tri-module data augmentation framework. We begin with the baseline model to reconstruct 3D human Gaussians from RGB frames (see Section~\ref{baseline}). The Generator module then creates unseen pose and camera variations by sampling Gaussian perturbations, producing coarse images (refer to Section~\ref{generator}). Using this curated training data (introduced in Section~\ref{data_construction}), the Refiner module transforms these coarse images into photorealistic representations, with details in Section~\ref{refiner}. Finally, the Examiner evaluates the refined images to identify high-quality pseudo GTs, as described in Section~\ref{examiner}.

\subsubsection{\textbf{Baseline Method}} \label{baseline}

Our baseline model is primarily based on ExAvatar~\cite{moon2024exavatar}, with specific operations illustrated in the lower left corner of Figure~\ref{fig: pipeline}. It employs a typical framework for learning 3D human avatars from monocular video, using a trainable triplane to capture geometric features. These features are queried at each vertex of a canonical SMPL mesh and fed into learnable MLPs to predict attributes of Gaussian points, thereby creating canonical 3D human Gaussians. To train the triplane and MLPs, these 3D Gaussian points are bonded to the vertices of the canonical SMPL mesh, and LBS is utilized to map Gaussian points into the observation space. Concretely, a matrix transformation is then applied to these space points based on the fitted SMPL pose ${\theta}$ derived from a video frame, followed by 3D Gaussian rendering under the camera pose $\pi$ that matches the frame view. The video frame serves as the GT for similarity calculations with the rendered image, providing supervision for gradients.

\subsubsection{\textbf{Generator}} \label{generator}
While the baseline model successfully reconstructs 3D human Gaussians from RGB frames, it struggles with the limited diversity of the training data, as monocular videos capture a narrow range of movements and views. To tackle this, the Generator synthesizes diverse unseen images by expanding the distribution of pose and camera parameters beyond the original video. It uses Gaussian distributions to sample perturbations, driving the baseline-reconstructed 3D Gaussians to generate coarse augmented images. The Generator generates coarse augmented images that reflect these novel variations, thereby enriching the training data and laying the groundwork for improved model ability.


In the baseline method outlined in Section~\ref{baseline}, we input the fitted SMPL pose $\theta$ to transform the canonical human 3D Gaussians $\mathcal{G}_c$ into observation 3D Gaussians $\mathcal{G}_o$ for rendering. By adjusting the SMPL pose to create a new one, we can generate a human 3D Gaussian in a different observation pose. If the body pose is defined by $n$ parameters ($\theta \in \mathbb{R}^{n}$), each parameter influences the spatial configuration of body parts~\cite{SMPL-X:2019}. Using the fitted SMPL Pose $\theta$ as a reference, increasing or decreasing these parameters results in corresponding changes in the observation 3D Gaussians. We associate each parameter fluctuation $\Delta \theta_i$ with a Gaussian distribution $N_i(\mu, \Sigma)$ for sampling. Larger sampled values yield greater differences between the generated $\mathcal{G}_{novel}$ and the original $\mathcal{G}_o$ for the respective body parts. This sampling process allows us to obtain multiple $\mathcal{G}_{novel}$s, which can then be used to render images $I_c$s of diverse human poses using Gaussian rendering techniques~\cite{3DGaussian}.

Similarly, for camera pose $\pi$, we can render $\mathcal{G}_{novel}$ from views different from the original one $\pi$. We associate Gaussian distributions with the camera's angle, distance parameters, to sample adjustments to these parameters, obtaining $\Delta \pi$, thereby obtaining different rendering viewpoints.

\begin{figure}
    \centering
    \includegraphics[width=1.0\linewidth]{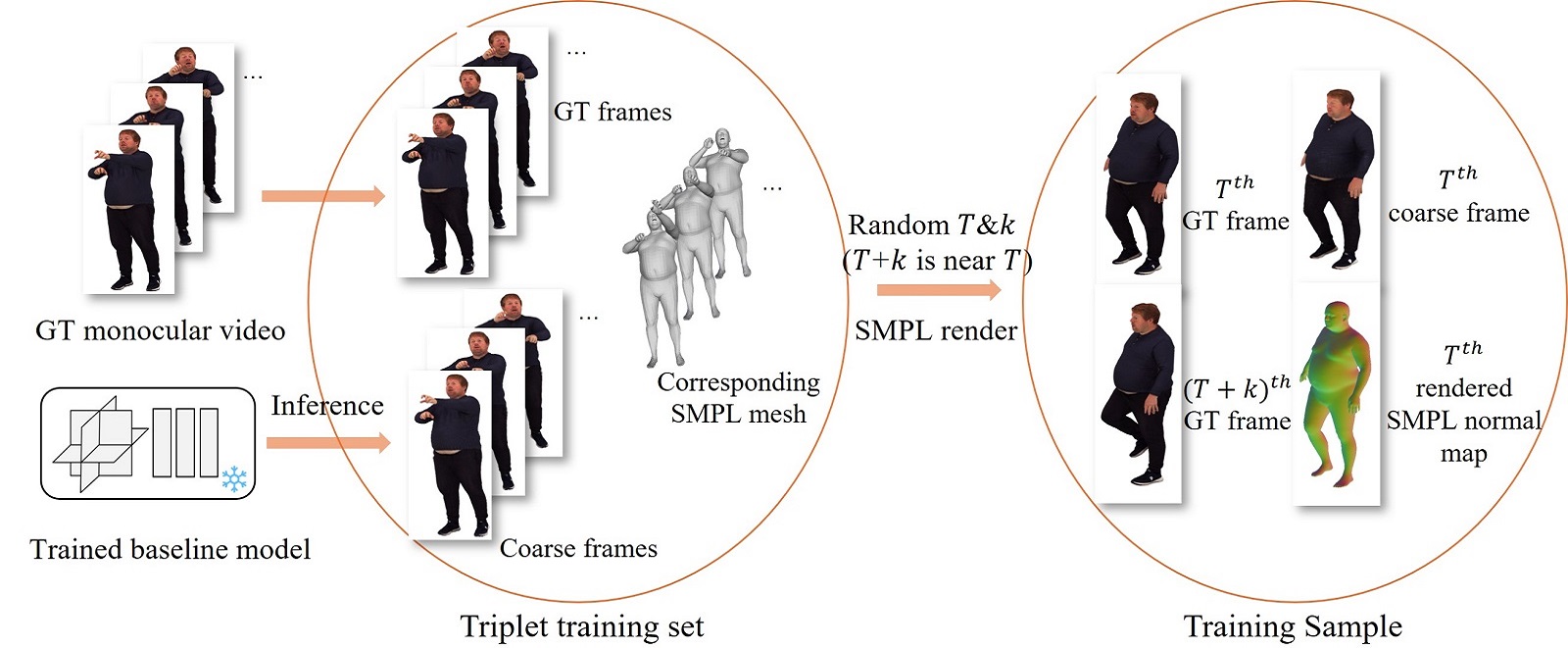}
    \vspace{-0.5cm}
    \caption{\textbf{Data construction for Refiner and Examiner.} We infer coarse frames from the trained baseline model using all the SMPL and camera parameters corresponding to GT frames in the monocular video. Along with the SMPL meshes mapped from the SMPL parameters, we compose a triplet set. We extract the GT and coarse frames at time $T$ and render the SMPL normal map, along with the GT frame at time $T+k$, to form a training sample.}
    \label{fig: data_construct}

\end{figure}

\subsubsection{\textbf{Data Construction}} \label{data_construction}



Using the aforementioned Generator, we create new images $I_c$ from an original sample (consisting of a GT frame $I$, its fitted SMPL pose $\theta$, and camera pose $\pi$). This is achieved by generating a new SMPL pose ${\theta}+\Delta \theta$ and adjusted camera parameters $\pi + \Delta \pi$. However, the resulting images are often coarse due to the limitations of the basic reconstruction model, making them not ideal for training data. To enhance the quality of these coarse images, we propose training a Refiner module. Before formally introducing the Refiner, we briefly outline the training data used for the training of the Refiner in this section for clarity.

Regarding the construction of training data, as illustrated in Figure~\ref{fig: data_construct}, we extract the corresponding SMPL poses and camera parameters for all monocular video frames about a subject~\cite{shen2023xavatar}. By feeding these corresponding SMPL poses and camera parameters into the trained baseline model for inference, we can obtain coarse frames that correspond one-to-one with the original monocular video frames. Meanwhile, we input the corresponding SMPL parameters into the SMPL model to obtain the SMPL meshes. By combining these three elements, we establish a triplet training set. 

To collect a training sample, we select a time $T$ and randomly pick a nearby time $T+k$. This gives us four items: the GT frame $I^{T}$and coarse frame $I_c^{T}$ at time $T$, the GT frame $I^{T+k}$ at time $T+k$, and the SMPL normal map $I_g^{T}$ rendered from the SMPL mesh at time $T$.

\begin{figure}
    \centering
    \includegraphics[width=1.0\linewidth]{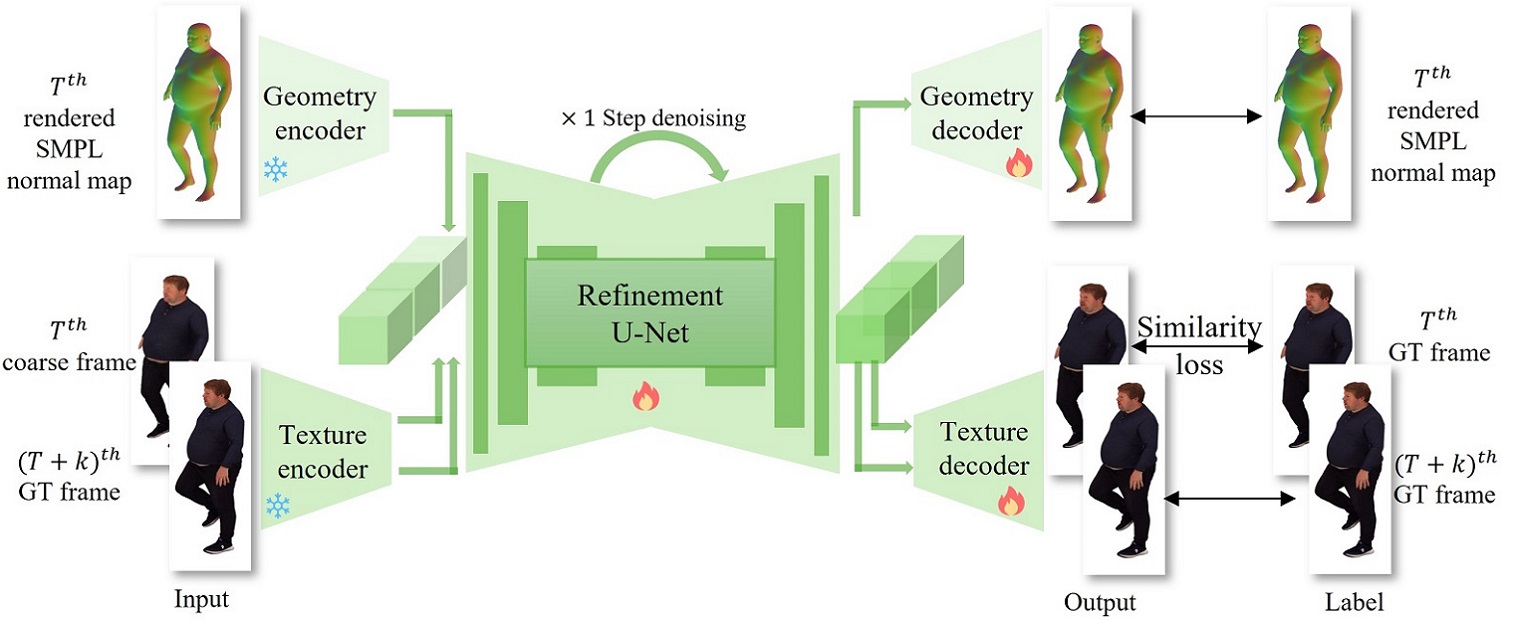}
    \vspace{-0.5cm}
    \caption{\textbf{Refiner module.} We take the refinement of the coarse frame as a single-step denoising process by treating it as a noisy image and using the GT frame as the original image, we utilize the corresponding SMPL as geometry conditions for the coarse frame, while the adjacent GT frame are used as texture condition. The model is trained to reconstruct the GT image.}
    \label{fig: refiner}
\end{figure}

\subsubsection{\textbf{Refiner}} \label{refiner}

As discussed in Sections~\ref{generator}, the samples generated by the generator offer the baseline model new images from various poses and views for training. Supplementing these additional data and distributions helps to more comprehensively optimize the model's training parameters. However, essentially, this kind of enhancement alone cannot provide the model with additional knowledge, which restricts further performance improvement. To address this limitation, we propose the Refinement module.


\myparagraph{Training} Figure~\ref{fig: refiner} shows the training process for the Refiner, using training samples in Section~\ref{data_construction}. Initially, two images (coarse $I_c^{T}$ and GT $I^{T+k}$), along with a normal map $I_g^{T}$, are encoded via VAE~\cite{SDTurbo} encoders $\mathcal{E}_t$ and $\mathcal{E}_g$, functioning as texture and geometric encoders. This encoding generates latent features $z_c = \mathcal{E}_t(I_c^{T})$, $z_t = \mathcal{E}_t(I^{T+k})$, and $z_g = \mathcal{E}_g(I_g^{T})$, each with dimensions $c*h*w$, where $c$ denotes the number of channels. These features are subsequently stacked to form a tensor with dimensions $3*c*h*w$.

The features, derived from two modalities and two views, are input into a U-Net~\cite{wu2025difix3d}, which contains self-attention layers that can facilitate interaction among the three-source features. Initially, these features are reshaped into dimensions $c*(3hw)$ for processing, enabling relevant texture and geometry features from $z_t$ and $z_g$ to attend to the coarse image features $z_c$. This allows $z_c$ to effectively absorb geometric information from $z_g$ and texture details from $z_t$, compensating for the degraded information.


After passing through the U-Net, the resulting latent features, $z_c^{'}$, $z_t^{'}$, and $z_g^{'}$, are decoded through the texture and geometric decoders to produce the refined images $I_c^{'}$, $I_t^{'}$, and $I_g^{'}$. These images are supervised using the GT images $I^{T}$ and $I^{T+k}$, along with the normal map $I_g^{T}$. The gradients are computed using L1, LPIPS~\cite{lpips}, and SSIM~\cite{wang2006modern_ssim} losses.

In this process, $I_c^{T}$ can be interpreted as the result of adding noise $\epsilon$ to $I^{T}$. By utilizing the conditional GT image $I^{T+k}$ and the normal map $I_g^{T}$, the training objective is to compensate for this noise $\epsilon$, thereby achieving refinement.

\myparagraph{Inference} As shown in Figure~\ref{fig: pipeline}, this Refiner, following Generator, takes the coarse frame $I_c$ as input, conditioned on its source GT frame $I$ (whose SMPL and camera poses are $\theta$ and $\pi$), and its corresponding SMPL normal map $I_g$ (We map the new SMPL pose ${\theta}+\Delta \theta$ to the SMPL mesh and render the normal map under the camera $\pi+\Delta \pi$.) to obtain the refined frame $I_r$. 

Since the model we train adopts one-step denoising, it own fast inference speed. We directly integrated its inference into the baseline model's one-time optimization process, enabling online refinement of the coarse frame.

\begin{figure}
    \centering
    \includegraphics[width=1.0\linewidth]{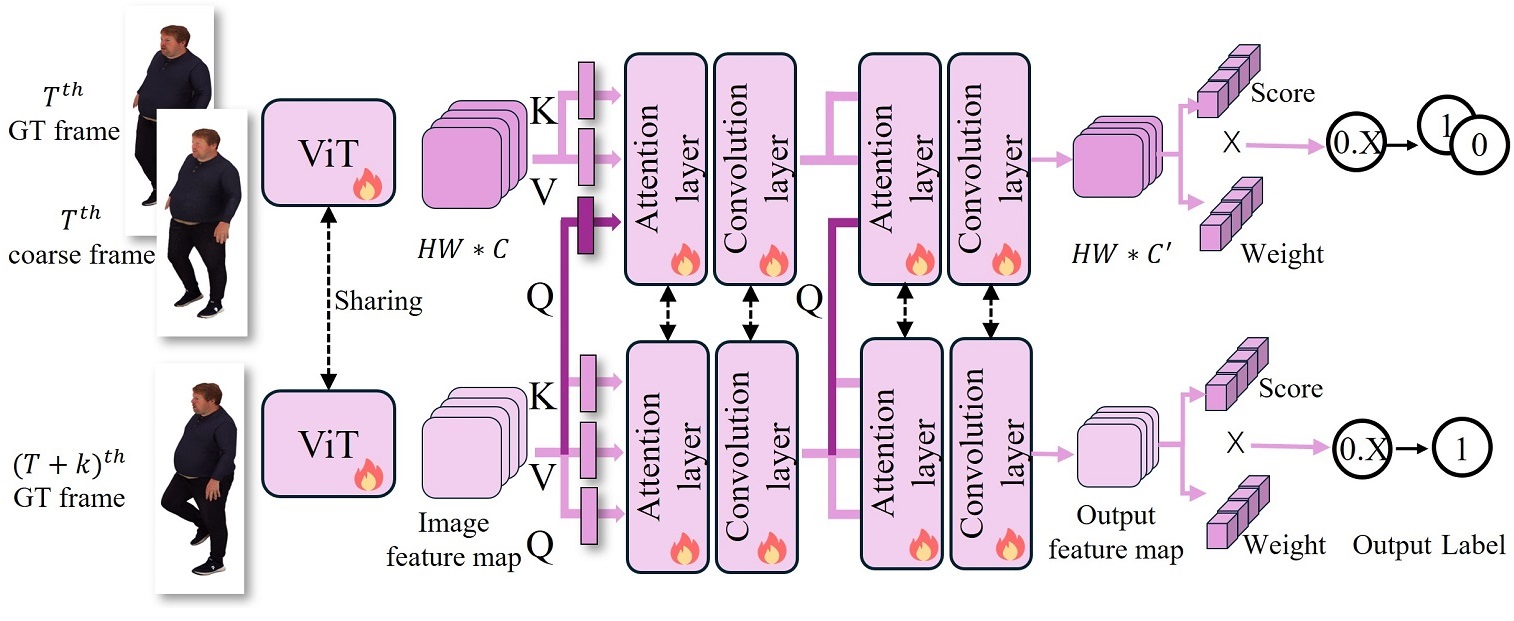}
        		\vspace{-0.5cm}
    \caption{\textbf{Examiner module.} We design a dual-branch similarity examination network. It samples the coarse frame $I_c^{T}$ or the GT frame $I^{T}$, and pair it with the GT frame $I^{T+k}$ to form a tuple. In the lower branch, we directly input $I^{T+k}$. We extracts features from $I^{T+k}$ to serve as queries, which interact with the keys and values derived from $I_c^{T}$ in the upper branch.}
    \label{fig: examiner}
\end{figure}

\begin{table*}
\begin{center}

    \caption{\textbf{Comparisons of 3D human avatars on the test set of X-Humans dataset}. Methods with $*$ use additional depth maps for the training.}\label{sota_xhuman}

\scalebox{1}{
\begin{tabular}
{l|ccc|ccc|ccc}
\toprule

\multirow{2}{*}{Methods}  & \multicolumn{3}{c}{00028} & \multicolumn{3}{c}{00034} & \multicolumn{3}{c}{00087}  \\
& PSNR $\uparrow$ & SSIM $\uparrow$ & LPIPS $\downarrow$  & PSNR $\uparrow$ & SSIM $\uparrow$ & LPIPS $\downarrow$  & PSNR $\uparrow$ & SSIM $\uparrow$ & LPIPS $\downarrow$ \\

\midrule
     X-Avatar*~\cite{shen2023xavatar}  & $28.57$ & $0.976$ & $0.026$ & $28.05$ & $0.965$ & $0.035$ & $30.89$ & $0.970$ & $0.030$ \\  
ExAvatar~\cite{moon2024exavatar} & $30.58$ & $0.981$ &  $0.018$ & $28.75$ & $0.966$ & $0.029$ & $32.01$ & $0.972$ & $0.025$  \\ 
Ours & $\textbf{33.04}$ & $\textbf{0.985}$ & $\textbf{0.014}$ & $\textbf{29.87}$ & $\textbf{0.968}$ & $\textbf{0.026}$ & $\textbf{32.82}$ & $\textbf{0.974}$ & $\textbf{0.023}$\\    

\bottomrule
    \end{tabular}
}


\end{center}
\end{table*}

\begin{table}
\begin{center}

\caption{\textbf{Comparisons on the test set of NeuMan}.\label{exp_abl}}
\label{sota_neu}

\scalebox{1}{
\begin{tabular}
{l|ccc}
\toprule

Methods & PSNR $\uparrow$ & SSIM $\uparrow$ & LPIPS $\downarrow$  \\

\midrule
HumanNeRF~\cite{weng_humannerf_2022_cvpr} & 27.06 & 0.967 & 0.019 \\  
InstantAvatar~\cite{jiang2022instantavatar}  & 28.47 & 0.972 & 0.028 \\  
NeuMan~\cite{jiang2022instantavatar} & 29.32 & 0.972 & 0.014 \\ 
Vid2Avatar~\cite{guo2023vid2avatar} & 30.70 & 0.980 & 0.014\\  
3DGS-Avatar~\cite{3DGS-Avatar} & 28.99 & 0.974 & 0.016\\  
GaussianAvatar~\cite{hu2024gaussianavatar} & 29.94 & 0.980 & 0.012\\    
PGHM~\cite{peng2025parametric} & 31.85 & 0.987 & 0.017 \\
EVA~\cite{hu2024expressive_eva} & 32.02 & 0.982 & 0.015 \\
Vid2Avatar-Pro~\cite{guo2025vid2avatar} &32.71 & 0.983 & 0.012 \\
MonoCloth~\cite{guo2025vid2avatar} &33.53 & 0.986 & 0.012 \\
PriorAvatar~\cite{PriorAvatar} &33.64 & 0.984 & 0.011 \\
ExAvatar~\cite{moon2024exavatar} & {34.80} & 0.984 & 0.009 \\
Ours & \textbf{35.42} & \textbf{0.987} & \textbf{0.009} \\    

\bottomrule
    \end{tabular}
}


\end{center}
\end{table}

\subsubsection{\textbf{Examiner}} \label{examiner}

Using the Refiner~\ref{refiner}, we can enhance the quality of coarse frames with a diffusion network. However, due to the inherent randomness in the diffusion model, not all refined frames may accurately match the texture of real frames in the captured monocular video. To tackle this issue and ensure the authenticity of the refined images, we introduce the Examiner module. Its primary role is to evaluate the similarity between the refined frame and the GT frame, deciding whether to accept the refined frame as pseudo-GT.

\myparagraph{Training} As depicted in Figure~\ref{fig: examiner}, we start by sampling a coarse frame $I_c^{T}$ and a GT frame $I^{T+k}$ from our constructed dataset~\ref{data_construction}. These frames are then processed through Vision Transformer (ViT~\cite{dosovitskiy2020image_vit}) to extract image features, denoted as $f_c = V(I_c^{T})$ and $f_k = V(I^{T+k})$. $f_c, f_k \in \mathbb{R}^{C*(HW)}$, where $(HW)$ is the number of spatial patches.

In the lower branch, $f_k$ is directly processed through the attention layer~\cite{vaswani2017attention}. This involves three linear layers to generate $K(f_k)$, $Q(f_k)$, and $V(f_k)$. An attention operation is then performed, denoted as $Attn_1(K(f_k), Q(f_k), V(f_k))$, during which we record $Q(f_k)$. In contrast, the processing of $f_c$ is slightly different. We pass $f_c$ through the same attention layer but only derive $K(f_c)$ and $V(f_c)$ via linear layers. They are combined with the recorded $Q(f_k)$ to execute the attention operation $Attn_1(K(f_c), V(f_c), Q(f_k))$.

This operation uses the features $f_c$ as both the key and the value, while $f_k$ acts as the query. This configuration is designed to examine the semantic similarity between different channels in the feature map~\cite{lee2024task_channel, 2020Thedevil, hu2018squeeze}. Here, we use the real image semantics as a reference to emphasize those genuine semantic features in $f_c$ related to the real image.

The output from the attention layer is reshaped to $C*H*W$ dimensions, followed by a convolution layer to squeeze the channels, promoting semantic interaction~\cite{hu2018squeeze}. This process integrates detailed semantics, resulting in $f_c^{1}$ and $f_k^{1} \in \mathbb{R}^{HW*C^1}$. They are passed and processed through successive blocks in the same manner, as shown in Figure~\ref{fig: examiner}.


Ultimately, we obtain ${f_c^{n}}$ and ${f_k^{n}} \in \mathbb{R}^{HW*C^n}$. Each is processed through two learnable MLPs to yield ${f_c^{s}}$ and ${f_c^{w}} \in \mathbb{R}^{HW}$, assigning scores and weights to patches. They are aggregated into the final score via weighted average. As shown in Figure~\ref{fig: examiner}, for the branch with $I^{T+k}$ as the sole input, the expected training output similarity score is 1. Conversely, for the interaction between $I_c^{T}$ and $I^{T+k}$, the expected score is 0. Moreover, during training, we not only sample $I_c^{T}$ into the interaction branch but also real frame $I^{T}$ with equal probability to compute similarity against $I^{T+k}$, where the expected output score is 1.


\myparagraph{Inference} It can refer to Figure~\ref{fig: pipeline}. In practical use, we obtain two refined images $I_r$s from an original sample ($I$, $\theta$, $\pi$) via the Generator and Refiner. Both of them will be compared to $I$ by the Examiner to calculate similarity, and we take the one with the higher similarity as the pseudo GT and discard the other one. This inference is also integrated into the baseline model’s one-time optimization.


\section{Experiments}
\label{sec:exp}


\myparagraph{Datasets and Evaluation Metrics} 
To compare our method with existing methods, we follow the protocols established by previous work, ExAvatar~\cite{moon2024exavatar}. We adopt the open monocular video dataset, Xhuman~\cite{shen2023xavatar}, focusing on subjects 00028, 00034, and 00087, alongside the NeuMan~\cite{jiang2022neuman} dataset, maintaining the same train/test splits as in prior work. For training our Refiner and Examiner, we constructed our dataset using the Xhuman dataset~\cite{shen2023xavatar}, which includes 20 human subjects. We excluded the aforementioned three subjects and used the remaining 17 for our training. To assess the quality of our reconstruction, we adopt evaluation metrics from ExAvatar. Specifically, we computed \textbf{PSNR}, \textbf{SSIM}~\cite{wang2006modern_ssim}, and \textbf{LPIPS}~\cite{lpips} for our results in comparison to GT. More details of datasets, implementation and more experiments can be found in the \textbf{supplementary materials}.


\begin{table}
\begin{center}

\caption{\textbf{Module ablation study.} Over the baseline, we gradually add the three proposed modules, and the model's performance improves progressively as we incorporate each new module.\label{exp_abl_module}}

\scalebox{1}{
\begin{tabular}
{l|ccc}
\toprule

\multirow{2}{*}{Methods}  & \multicolumn{3}{c}{00028}  \\
& PSNR $\uparrow$ & SSIM $\uparrow$ & LPIPS $\downarrow$  \\

\midrule
     Baseline  & 30.5842 & 0.9814 & 0.0181 \\  
+ Generator & 31.2905 & 0.9825 & 0.0170 \\ 
+ Generator\&Refiner & 32.6586 & 0.9847 & 0.0146\\    
+ Generator\&Refiner\&Examiner & \textbf{33.0412} & \textbf{0.9851} & \textbf{0.0143} \\    

\midrule

& \multicolumn{3}{c}{00034} \\

\midrule
     Baseline   & 28.7533 & 0.9659 &  0.0292\\  
+ Generator & 28.9654  & 0.9665 &  0.0280\\ 
+ Generator\&Refiner & 29.6413 & 0.9678 & 0.0263 \\    
+ Generator\&Refiner\&Examiner & \textbf{29.8745} & \textbf{0.9681} & \textbf{0.0261}\\    
\bottomrule
    \end{tabular}
}


\end{center}
\end{table}

\begin{table}[!t]
\begin{center}

\caption{\textbf{More ablation study about Refiner.} Based on the Generator, we explore the impact of different conditions in the Refiner on the final results. Without any guiding conditions, the refinement leads to a slight decrease. However, as we add texture and geometry conditions, the results gradually improve. \label{exp_abl_refiner}}

\scalebox{0.95}{
\begin{tabular}
{l|ccc}
\toprule

\multirow{2}{*}{Methods}  & \multicolumn{3}{c}{00028}  \\
& PSNR $\uparrow$ & SSIM $\uparrow$ & LPIPS $\downarrow$  \\

\midrule
Coarse image Generation & 31.2905 & 0.9825 & 0.0170 \\ 
+ Refinement (Without condition)  & 30.9897 & 0.9818 & 0.0172\\    
+ Refinement (Only Tex. condition) & 32.1369 & 0.9841 & 0.0155\\    
+ Refinement (Tex.\&Geo. condition) & \textbf{32.6586} & \textbf{0.9847} & \textbf{0.0146}\\    

\midrule

& \multicolumn{3}{c}{00034} \\

\midrule
Coarse image Generation & 28.9654  & 0.9665 &  0.0280\\ 
+ Refinement (Without condition) & 28.9375 & 0.9663 & 0.0283\\    
+ Refinement (Only Tex. condition) & 29.4364 & 0.9674 & 0.0266\\    
+ Refinement (Tex.\&Geo. condition) & \textbf{29.6413} & \textbf{0.9678} & \textbf{0.0263} \\    
\bottomrule
    \end{tabular}
}

\end{center}
\end{table}

\subsection{Quantitative Comparison with SOTA methods}

In Tables~\ref{sota_xhuman} and~\ref{sota_neu}, our method outperforms existing approaches on both X-Humans and NeuMan datasets. Specifically, on X-Humans, for the subject 00028, our method outperforms the baseline Exavatar by $2.46$ in PSNR and $0.004$ in SSIM. On NeuMan, our method yields a PSNR improvement of $0.62$. These improvements stem from our tri-module augmentation framework, which enriches training data diversity via the Generator, refines synthetic-real consistency through the Refiner, and ensures high-quality samples via the Examiner, collectively boosting reconstruction results. \textbf{The results of other methods in the tables mainly come from ExAvatar~\cite{moon2024exavatar}. We obtain their respective results from works ~\cite{peng2025parametric, guo2025vid2avatar, jin2026monocloth, PriorAvatar}. Moreover, we reproduce EVA~\cite{hu2024expressive_eva} for the results.}

\subsection{Quantitative Ablation Study}

\myparagraph{Module Ablation} We perform ablation studies on our modules (Generator, Refiner, Examiner) in Table~\ref{exp_abl_module}, observing consistent gains with each addition. The reason lies in that the Generator synthesizes diverse unseen images via Gaussian-guided pose/camera perturbations. Supplementing these additional data and distributions helps more comprehensively optimize the model’s training parameters. The Refiner uses diffusion networks to refine the Generator’s coarse frames, boosting image quality, thus improving reconstruction results. The Examiner filters refined frames by GT similarity, retaining only high-consistency pseudo-GT.

\begin{figure*}
    \centering
    \setlength{\belowcaptionskip}{-0.1cm}
    \includegraphics[width=1\linewidth]{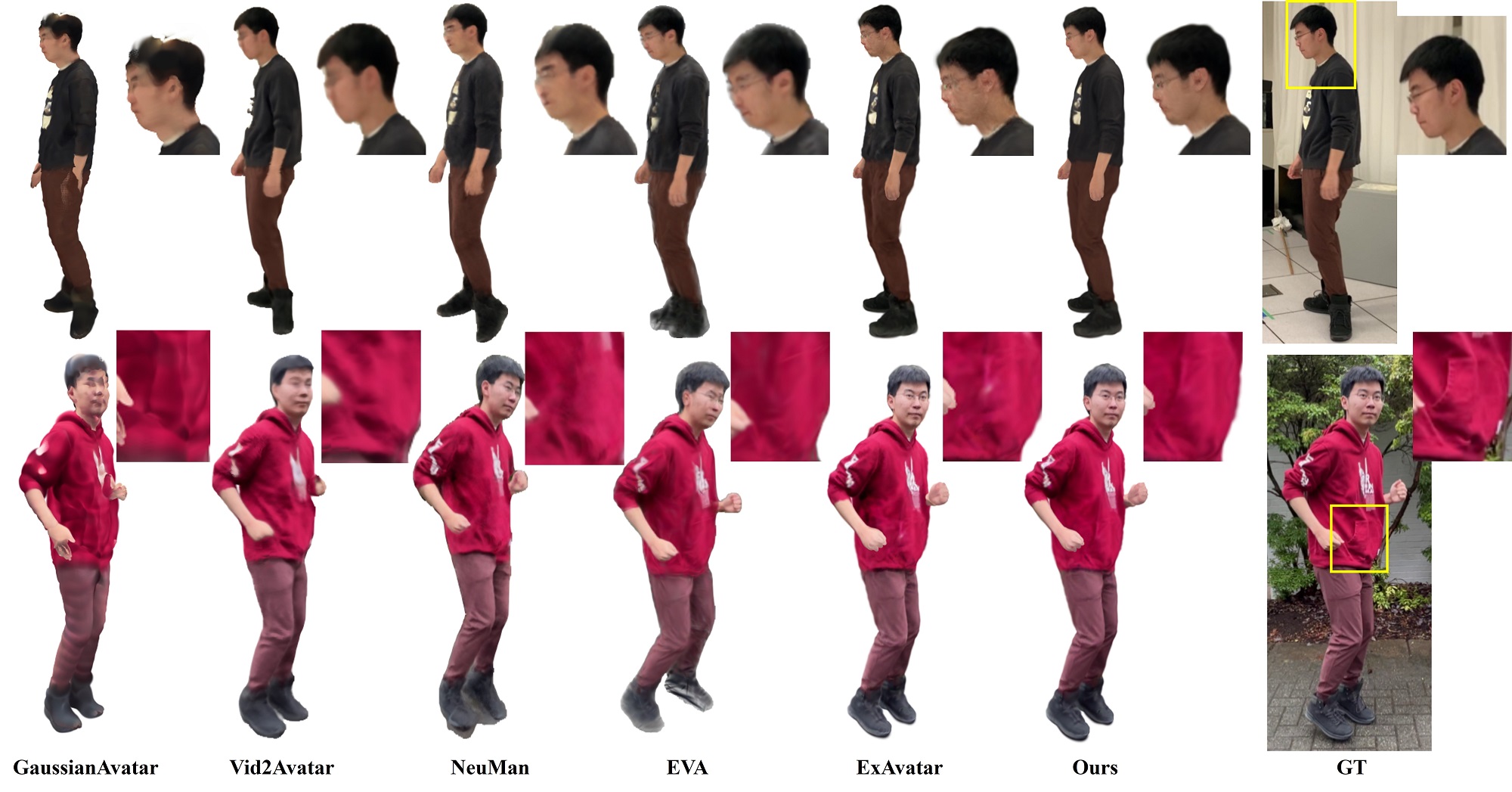}
    		\vspace{-0.4cm}

    \caption{\textbf{Visual comparison with SOTA methods on Neuman.} Compared to current methods, our approach can achieve better reconstruction performance, such as reducing artifacts on side profiles of faces and providing clearer details of clothing pockets.}
    \label{fig:vis_comp_neuman}

\end{figure*}

\begin{figure}
    \centering
    \setlength{\belowcaptionskip}{-0.1cm}
    \includegraphics[width=1\linewidth]{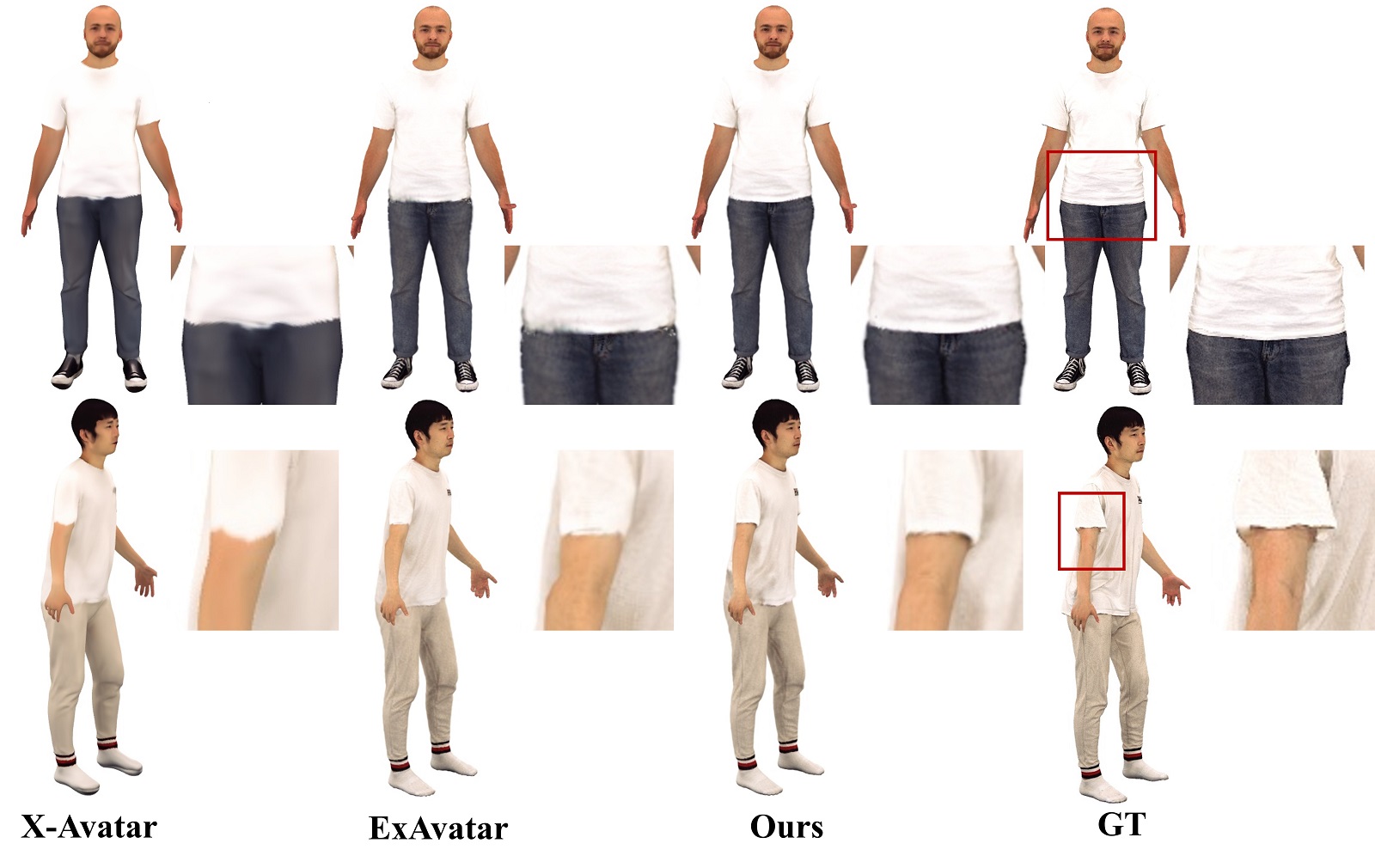}
    		\vspace{-0.5cm}

    \caption{\textbf{Visual comparison with SOTA methods on X-Human.} Compared to these existing approaches, our method delivers superior reconstruction performance, particularly in capturing details like clothing wrinkles and edges.}
    \label{fig:vis_comp_xhuman}

\end{figure}

\myparagraph{Refiner Ablation} We ablate the Refiner’s conditioning factors in Table~\ref{exp_abl_refiner}. Unconditional refinement degrades performance, as it can randomly alter critical details, as illustrated in Figure~\ref{fig:abl_why_wotex_drop}. In contrast, incorporating texture and geometry conditions leads to consistent performance improvements. Texture conditions ensure detail similarity to the original frame, while geometry conditions maintain geometric correctness. This is visually demonstrated in Figures~\ref{fig:abl_why_wotex_drop} and~\ref{fig:abl_geo_cond}.

\subsection{Visualiztion}

\begin{figure*}[]
    \centering
    \setlength{\belowcaptionskip}{-0.1cm}
    \includegraphics[width=1\linewidth]{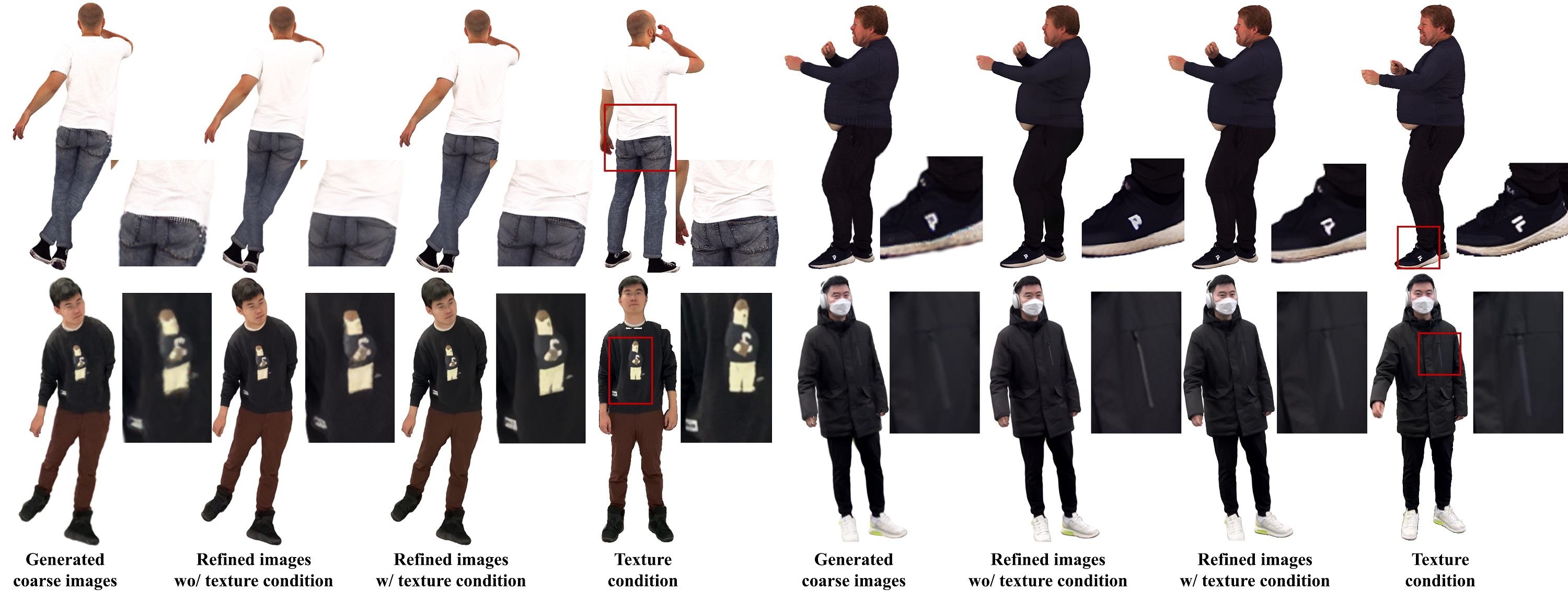}
    		\vspace{-0.7cm}

    \caption{\textbf{Visual ablation of Refiner.} We show the refinement effects of the Refiner module before and after incorporating texture conditions, highlighting a notable improvement in image quality. The introduction of texture conditions enhances details like wrinkles and patterns on clothing in the left two groups. In contrast, without these conditions, the Refiner would incorrectly enhance certain features, leading to false hallucination, as shown in the right two groups with shoe patterns and clothing designs.}
    \label{fig:abl_why_wotex_drop}

\end{figure*}

\myparagraph{Comparison with SOTA Methods} In Figures~\ref{fig:tease}, ~\ref{fig:vis_comp_neuman}, and ~\ref{fig:vis_comp_xhuman}, we demonstrate that our method outperforms existing approaches in detail reconstruction, particularly for clothing wrinkles and edges. This superiority arises from our framework's supplement of high-quality data: the Generator enhances training diversity with varied fabric textures and wrinkle configurations, the Refiner boosts texture precision, and the Examiner ensures data authenticity. Together, these components equip the model with richer cues for capturing clothing nuances, unlike baseline methods constrained by limited training data. \textbf{Notably, some works ~\cite{peng2025parametric, guo2025vid2avatar, jin2026monocloth, PriorAvatar} have not open-sourced their codes.}

\begin{figure}
    \centering
    \setlength{\belowcaptionskip}{-0.1cm}
    \includegraphics[width=1\linewidth]{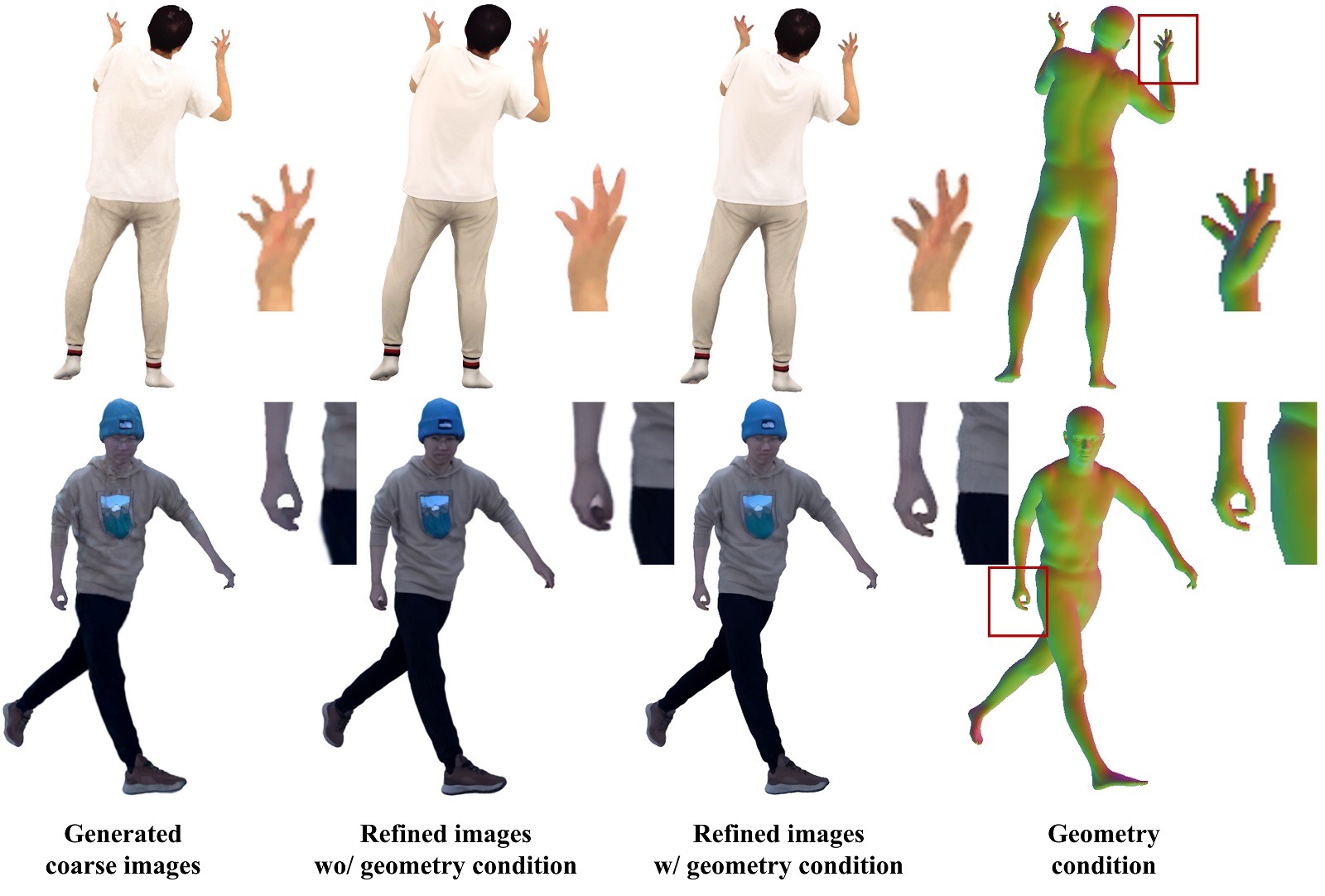}
    		\vspace{-0.6cm}

    \caption{\textbf{Visual ablation about the geometry condition in Refiner.}  It is observed that introducing geometric conditions in the Refiner module enhances the improvement of human geometric quality, such as the details of the fingers.}
    \label{fig:abl_geo_cond}

\end{figure}

\myparagraph{Visual Ablation} Figures~\ref{fig:abl_why_wotex_drop} and~\ref{fig:abl_geo_cond} showcase the Refiner's effectiveness in refining coarse images.  We find that incorporating texture and geometric conditions enhances performance: texture condition reduces the model's tendency to produce artifacts, like the pattern in the clothes, while adding geometric conditions improves the model's ability to enhance human geometric details, like the human fingers. The reason is that during the use of the diffusion model, the model can randomly introduce some false illusions. However, by adding texture semantic constraints of the neighborhood frames and geometric constraints, the direction of model diffusion can be effectively corrected.


\begin{figure}
    \centering
    \setlength{\belowcaptionskip}{-0.1cm}
    \includegraphics[width=1\linewidth]{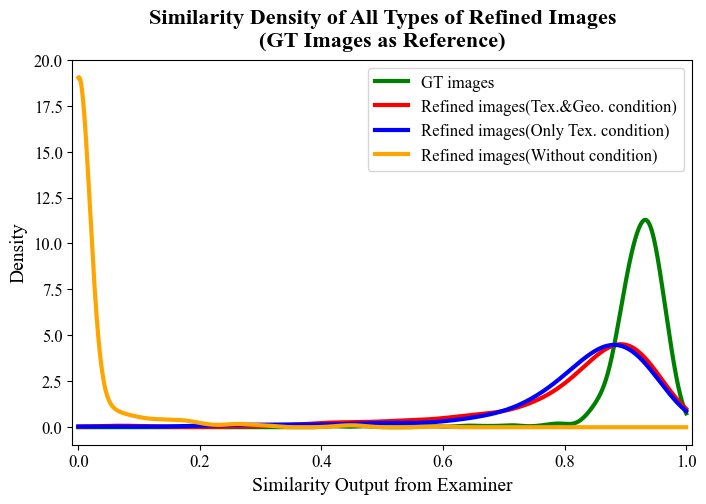}
    		\vspace{-0.2cm}

    \caption{\textbf{Visualization about Examiner.} It is found that Examiner can distinguish between different refined images. When GT images are input, the similarity values approach 1, while for images refined without conditions, they approach 0. Applying conditions like texture and geometry improves the authenticity of refined images, though some instances of low authenticity still occur.}
    \label{fig:line_tex_geo_cond}

\end{figure}

\myparagraph{Examiner Visualization} In Figure~\ref{fig:line_tex_geo_cond}, we present the similarity density functions of the Examiner’s outputs for various image types. The figure shows the Examiner’s ability to differentiate between high-quality refined frames (closely matching GT) and low-quality ones (affected by false hallucination, like the examples shown in Figure~\ref{fig:abl_why_wotex_drop}). This ability is vital because the Refiner, while enhancing frame quality through diffusion, can introduce randomness that leads to semantics-misaligned outputs. Without the Examiner, these flawed samples could contaminate training. By filtering out low-similarity frames, the Examiner ensures that more authentic refined images are retained. Due to the geometry condition's ability to assist with fine-grained geometric details (Figure~\ref{fig:abl_geo_cond}), it can further enhance similarity.

\section{Conclusion}
\label{sec:conclusion}
In this paper, we presented \projecttitle, a tri-module augmented framework that effectively addresses the data scarcity and viewpoint limitations inherent in monocular avatar reconstruction. By synergizing a distribution-perturbed Generator, a one-step diffusion-based Refiner, and a dual-branch attention Examiner, our approach systematically enriches training diversity while ensuring the photorealism and reliability of synthetic samples. Extensive experiments on the X-Humans and NeuMan benchmarks demonstrate that \projecttitle achieves state-of-the-art performance, greatly improving pose generalization and rendering fidelity. Ultimately, our work provides a robust solution for recovering high-fidelity, subject-specific 3D avatars from limited monocular input, paving the way for more accessible and immersive digital human applications.



\bibliographystyle{ACM-Reference-Format}
\bibliography{sample-base}









\end{document}